\documentclass[letterpaper]{article} 
\usepackage{aaai2026}  
\usepackage{times}  
\usepackage{helvet}  
\usepackage{courier}  
\usepackage[hyphens]{url}  
\usepackage{graphicx} 
\usepackage{natbib}  
\usepackage{caption} 
\usepackage{amsmath}
\usepackage{amssymb}
\usepackage{algorithm}
\usepackage{algorithmic}
\usepackage{bm}
\usepackage{svg}
\usepackage{multirow}
\usepackage{diagbox}
\usepackage{adjustbox}
\usepackage{tabularx}
\usepackage{rotating}
\usepackage{booktabs}
\usepackage{xcolor}
\usepackage{colortbl}
\usepackage{xr}
\usepackage{makecell}
\usepackage{pifont}
\definecolor{my_red}{RGB}{201, 132, 132}
\definecolor{my_orange}{RGB}{206, 129, 82}
\definecolor{my_yellow}{RGB}{193, 165, 123}
\definecolor{my_green}{RGB}{120, 166, 73}
\definecolor{my_cyan}{RGB}{106, 156, 137}
\definecolor{my_blue}{RGB}{109, 147, 180}
\definecolor{my_purple}{RGB}{133, 134, 200}
\definecolor{my_gray}{RGB}{124, 136, 148}

\usepackage{amsmath}
\usepackage{amssymb}
\usepackage{algorithm}
\usepackage{algorithmic}
\usepackage{bm}
\usepackage{svg}
\usepackage{multirow}
\usepackage{diagbox}
\usepackage{adjustbox}
\usepackage{tabularx}
\usepackage{rotating}
\usepackage{booktabs}
\usepackage{xcolor}
\usepackage{colortbl}
\usepackage{xr}
\usepackage{makecell}
\usepackage{pifont}
\usepackage{tcolorbox}
\definecolor{my_red}{RGB}{201, 132, 132}
\definecolor{my_orange}{RGB}{206, 129, 82}
\definecolor{my_yellow}{RGB}{193, 165, 123}
\definecolor{my_green}{RGB}{120, 166, 73}
\definecolor{my_cyan}{RGB}{106, 156, 137}
\definecolor{my_blue}{RGB}{109, 147, 180}
\definecolor{my_purple}{RGB}{133, 134, 200}
\definecolor{my_gray}{RGB}{124, 136, 148}

\usepackage{newfloat}
\usepackage{listings}

\usepackage{newfloat}
\usepackage{listings}
\DeclareCaptionStyle{ruled}{labelfont=normalfont,labelsep=colon,strut=off} 
\floatstyle{ruled}
\newfloat{listing}{tb}{lst}{}
\floatname{listing}{Listing}
\title{Laytrol: Preserving Pretrained Knowledge in Layout Control for Multimodal Diffusion Transformers}
\author{
    Sida Huang\textsuperscript{\rm 1,2},  
    Siqi Huang\textsuperscript{\rm 1,2},
    Ping Luo\textsuperscript{\rm 3},
    Hongyuan Zhang\textsuperscript{\rm 2, 3}\thanks{Corresponding author.}
}
\affiliations{
    \textsuperscript{\rm 1}School of Artificial Intelligence, OPtics and ElectroNics (iOPEN), \\
    Northwestern Polytechnical University, Xi’an 710072, P. R. China \\
    \textsuperscript{\rm 2}Institute of Artificial Intelligence (TeleAI), China Telecom\\
    \textsuperscript{\rm 3}The University of Hong Kong\\
    \{sidahuang2001, 4777huang\}@gmail.com, pluo@cs.hku.hk, hyzhang98@gmail.com
}

\usepackage{bibentry}

\begin{document}

\maketitle
\begin{figure*}[t]
    \centering
    \includegraphics[width=\linewidth]{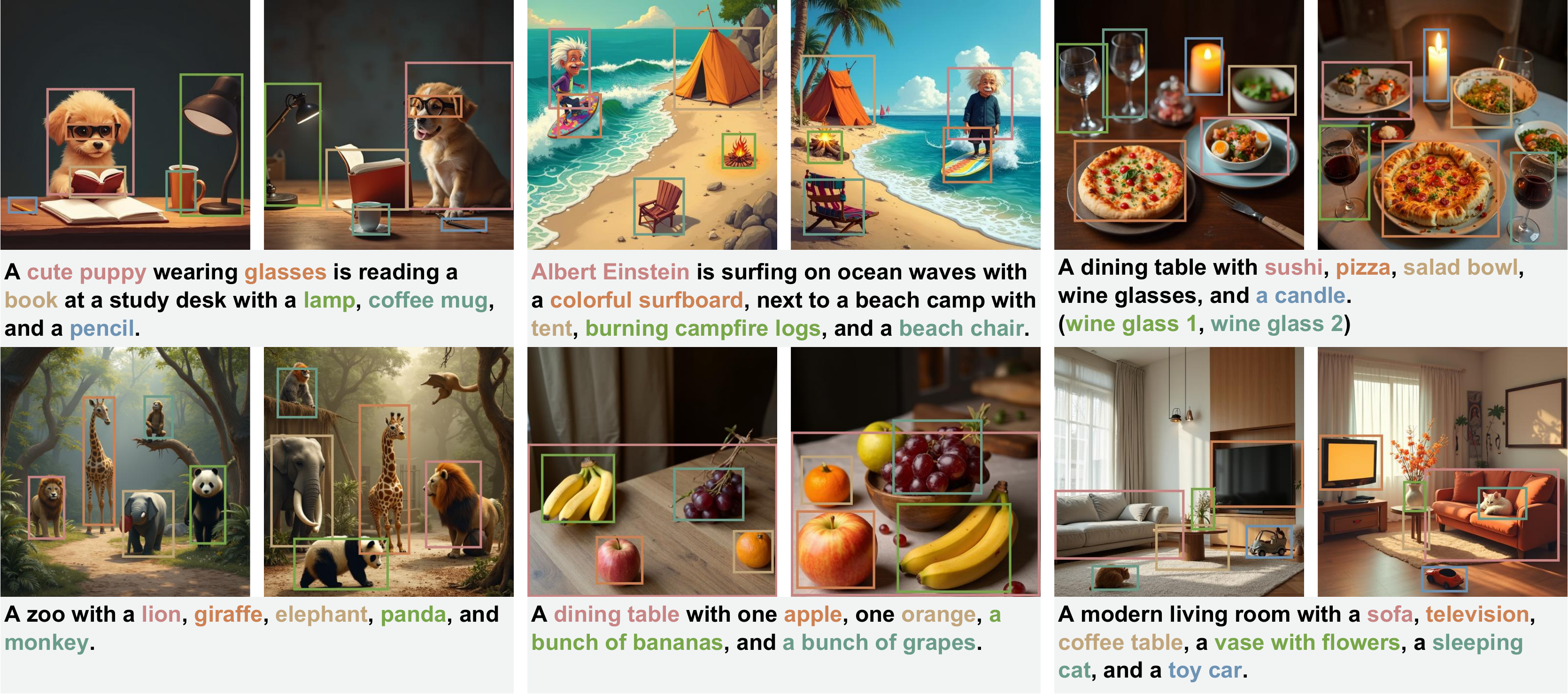}
    \caption
    {
        \textbf{Layout-to-image results of Laytrol}.
        The text in the same color as the bounding box corresponds to the local prompt.
        Laytrol enables layout-conditioned control in MM-DiT while effectively retaining the knowledge learned during the pre-training stage, and simultaneously mitigates the domain shift introduced by fine-tuning.
    }
    \label{fig_visual1}
\end{figure*}


\begin{abstract}
With the development of diffusion models, enhancing spatial controllability in text-to-image generation has become a vital challenge. As a representative task for addressing this challenge, layout-to-image generation aims to generate images that are spatially consistent with the given layout condition. Existing layout-to-image methods typically introduce the layout condition by integrating adapter modules into the base generative model. However, the generated images often exhibit low visual quality and stylistic inconsistency with the base model, indicating a loss of pretrained knowledge. To alleviate this issue, we construct the \textbf{Lay}out \textbf{Syn}thesis (LaySyn) dataset, which leverages images synthesized by the base model itself to mitigate the distribution shift from the pretraining data. Moreover, we propose the \textbf{Lay}out Con\textbf{trol} (Laytrol) Network, in which parameters are inherited from MM-DiT to preserve the pretrained knowledge of the base model. To effectively activate the copied parameters and avoid disturbance from unstable control conditions, we adopt a dedicated initialization scheme for Laytrol. In this scheme, the layout encoder is initialized as a pure text encoder to ensure that its output tokens remain within the data domain of MM-DiT. Meanwhile, the outputs of the layout control network are initialized to zero. In addition, we apply Object-level Rotary Position Embedding to the layout tokens to provide coarse positional information. Qualitative and quantitative experiments demonstrate the effectiveness of our method.
\end{abstract}

\begin{links}
    \link{Code}{https://github.com/HHHHStar/Laytrol}
\end{links}

\section{Introduction}
\label{sec_intro}

Diffusion models have significantly promoted the development of text-to-image (T2I) generation.
Among them, U-Net-based Stable Diffusion \cite{LDM} models have been widely used in the T2I community due to the efficiency and effectiveness.
Recently, MM-DiT introduced the transformer architecture into diffusion models, leading to the emergence of more advanced T2I models such as Stable Diffusion 3 \cite{SD3} and FLUX \cite{FLUX}.

To enhance spatial controllability in T2I models, the task of layout-to-image generation is proposed, which aims to generate different objects within specified regions of the image.
For this task, many adapter-based methods \cite{GLIGEN, MIGC, CreatiLayout} have been proposed. 
These methods insert new adapter modules into the base generative model and train the model on datasets with layout annotations.
However, we find that the images generated by adapter-based methods exhibit low visual quality and stylistic inconsistency with the base model, as illustrated in Figure \ref{fig_visual2}.
The factors contributing to this issue can be analyzed from two perspectives: the training dataset and the control module.
On the one hand, previous training datasets are primarily based on COCO \cite{COCO} or subsets of LAION \cite{laion-5b, laion-aesthetics-v2}. 
Distribution shift exists between these datasets and the pretraining data of base generative models.
On the other hand, control modules in existing methods are trained from scratch, which prevents them from effectively inheriting the pretrained knowledge of the base model.
Therefore, \textbf{\textit{we focus on effectively utilizing the pretrained knowledge of the base generative model to guide dataset construction and control module parameter initialization}}.

\textbf{Regarding dataset construction}, we first generate images using base generative model FLUX \cite{FLUX}, and then annotate their layouts using open-source models such as Grounding DINO \cite{GroundingDINO}.
Images generated by the model itself can effectively retain the image style and high-quality details derived from the pretraining knowledge, thereby mitigating the distribution shift from the pretraining data.
During this process, we observe that the model tends to produce images with repetitive layout patterns. To mitigate this layout bias, we propose layout prompting, which augments object description by randomly incorporating spatial and size-related phrases.

\textbf{Regarding initialization of control module parameters}, inspired by ControlNet \cite{controlnet}, we propose to incorporate copied parameters from MM-DiT into the corresponding layout control modules.
Leveraging copied parameters requires satisfying two initialization conditions prior to training:
\begin{itemize}
  \item[\textbf{\textit{C1}}] The input to the layout control modules must lie within their own data domain.
  \item[\textbf{\textit{C2}}] The output of the layout control modules must be initialized to zero.
\end{itemize}
At the start of training, \textit{C1} ensures that the copied parameters can be properly activated, and \textit{C2} ensures that the base generative network is not disturbed by unstable control conditions.
To satisfy \textit{C1}, we initialize the layout encoder to be functionally equivalent to the text encoder, for which its output tokens lie within the domain of MM-DiT.
In  the following training process, spatial information is gradually injected into the layout encoder.
To satisfy \textit{C2}, we add a zero-initialized linear layer, ensuring that the initial output of the layout control modules is zero.

Overall, our contributions can be summarized as follows:
\begin{itemize}
  \item We propose \textbf{Lay}out Con\textbf{trol} (Laytrol) Network, a layout-to-image generation method that preserves pretrained knowledge by leveraging parameter copying.
  \item We introduce \textbf{Lay}out \textbf{Syn}thesis (LaySyn) dataset, which leverages the base generative model FLUX for image synthesis to alleviate distribution shift. To mitigate layout bias in this process, we employ layout prompting that randomly injects layout phrases into object descriptions.
  \item We propose object-level Rotary Position Embedding (RoPE) to encoode coarse positional information for layout tokens.
\end{itemize}
\section{Related Works}
\subsection{Text-to-Image Generation}
Text-to-image(T2I) generation is the task of learning a conditional mapping from a natural-language description to a corresponding image. By leveraging noise \cite{pi-noise, VPN, PiNDA, PiNI, PiNGDA, MIN}, diffusion models \cite{DDPM, DDIM} have rapidly advanced and and have been applied across various domains \cite{adv-CPG, AVEdit, NFIG}. Open-source Stable Diffusion \cite{LDM, SDXL} have achieved promising performance in T2I synthesis. However, the limited representation capability of CLIP text encoder and U-Net architecture restricts both semantic comprehension and image quality. Recent studies\cite{Imagen, SD3}, employ T5 \cite{T5} as the text encoder to enhance the understanding ability of textual prompts. In addition, Diffusion Transformer (DiT) \cite{DiT} introduces the transformer architecture into the image generation domain, exhibiting superior scalability. Compared to diffusion models, rectified flow model \cite{rectified_flow} achieves improved performance by learning straight paths between two distribution points. Leveraging these advancements, Stable Diffusion 3 \cite{SD3} and FLUX \cite{FLUX} achieve state-of-the-art performance.

\subsection{Layout-to-Image Generation}
Since textual descriptions exhibit ambiguity in conveying spatial information, generating images that conform to specific spatial layouts has become increasingly important. Several training-free methods
\cite{multidiffusion, boxdiff} constrain object positions by optimizing noisy latents guided by attention maps. RPG \cite{RPG} utilizes a large language model (LLM) to design layouts, then generates each object individually and composes them in the latent space. These methods save training resources but typically require more inference steps. Moreover, they suffer from degraded image quality and lower layout fidelity. In contrast, training-based approaches such as GLIGEN \cite{GLIGEN} encode bounding box coordinates using Fourier embeddings \cite{nerf} and control layouts through a newly trained cross-attention layer. Similarly, MIGC \cite{MIGC} also propose dedicated layout control modules within the U-Net architecture.
Building on DiT, SiamLayout \cite{CreatiLayout} trains an MM-DiT-based control module for Stable Diffusion 3 \cite{SD3} and FLUX \cite{FLUX}.
However, these models train the newly inserted modules from scratch, lacking the utilization of pretrained knowledge.

\section{Method}

\label{sec_method}

\subsection{Preliminaries}

\subsubsection{Multimodal Diffusion Transformer}
To introduce the architecture of transformers in image generation, Multimodal Diffusion Transformer (MM-DiT) is proposed to jointly process image tokens $\mathbf{X} \in \mathbb{R}^{n \times d} $ and text tokens $\mathbf{C_T} \in \mathbb{R}^{m \times d}$. 
Analogous to a standard transformer module, these tokens are projected through linear layers to obtain $\mathbf{Q_I}, \mathbf{K_I}, \mathbf{V_I}$ and $\mathbf{Q_T}, \mathbf{K_T}, \mathbf{V_T}$, respectively.
Before computing attention, Rotary Position Embedding (RoPE) \cite{RoPE} is applied to the query $\mathbf{Q}$ and key $\mathbf{K}$ to encode relative position information. For a visual token $\mathbf{X}^{i,j}$ located at position $(i, j)$ in the 2D latent space, its corresponding query $\mathbf{Q_I}^{i,j}$ and key $\mathbf{K_I}^{i,j}$ are mapped as
\begin{equation} \label{eqn_rope}
    \begin{aligned}
        \mathbf{Q_I}^{i,j}=\mathbf{W_Q} \cdot \mathbf{X}^{i,j} \cdot \mathbf{R}(i,j);  \\
        \mathbf{K_I}^{i,j}=\mathbf{W_K} \cdot \mathbf{X}^{i,j} \cdot \mathbf{R}(i,j),
    \end{aligned}
\end{equation}
where $\mathbf{W_Q}$ and $\mathbf{W_K}$ are linear transformation matrices, $\mathbf{R}(i,j)$ denotes the rotation matrix. For text tokens, FLUX assigns a fixed position index of $(0,0)$, as their positional information has already been encoded during text embedding. These tokens are then concatenated as $\mathbf{Q} = \mathbf{Q_I} \oplus \mathbf{Q_T}$, $\mathbf{K} = \mathbf{K_I} \oplus \mathbf{K_T}$ and $\mathbf{V} = \mathbf{V_I} \oplus \mathbf{V_T}$. The concatenated tokens from both modalities are subsequently involved in the scaled dot-product attention computation:
\begin{equation} \label{eqn_attention}
        \text{Attention}(\mathbf{Q},\mathbf{K},\mathbf{V})=\text{Softmax}\left( \frac{\mathbf{Q} \mathbf{K}^\top}{\sqrt{d_k}} \right) \mathbf{V}.
\end{equation}
During computation, the product of two rotation matrices $\mathbf{R}(i,j)\cdot \mathbf{R}(i',j')^\top$ depends solely on their relative position $(i-i',j-j')$.

\subsubsection{Layout-to-Image Generation}
Layout-to-image generation aims to synthesize specified content for different regions of an image, thereby enhancing spatial controllability. 
The conditional input for this task consists of a global prompt $p_g$ and a layout condition $\mathcal{L}$, which comprises $N$ entities $\mathbf{e}$.
Each entity $\mathbf{e}_i$ is associated with a local prompt $p_i$ and a spatial position $\mathbf{b}_i$:
\begin{equation} \label{eqn_layout}
        \mathcal{L} = \{\mathbf{e}_1, ...,\mathbf{e}_N\}, \quad
        \text{where} \; \mathbf{e}_i = (p_i, \mathbf{b}_i).
\end{equation}
In the equation, $\mathbf{b}$ denotes the bounding box coordinates $(x_1,y_1,x_2,y_2)$.
To effectively represent the spatial positions of bounding boxes, GLIGEN \cite{GLIGEN} encodes these coordinates using Fourier embeddings \cite{nerf}.
Finally, the Fourier embeddings and the encoded prompts are transformed by a multi-layer perceptron (MLP) to obtain the final layout tokens $\mathbf{C_L}^i \in \mathbb{R}^d$:
\begin{equation} \label{eqn_GLIGEN}
        \mathbf{C_L}^i = \mathrm{MLP}(\mathrm{CLIP}(p_i)\oplus \mathrm{Fourier}(\mathbf{b}_i)).
\end{equation}

\begin{figure*}[t]
    \centering
    \includegraphics[width=\linewidth]{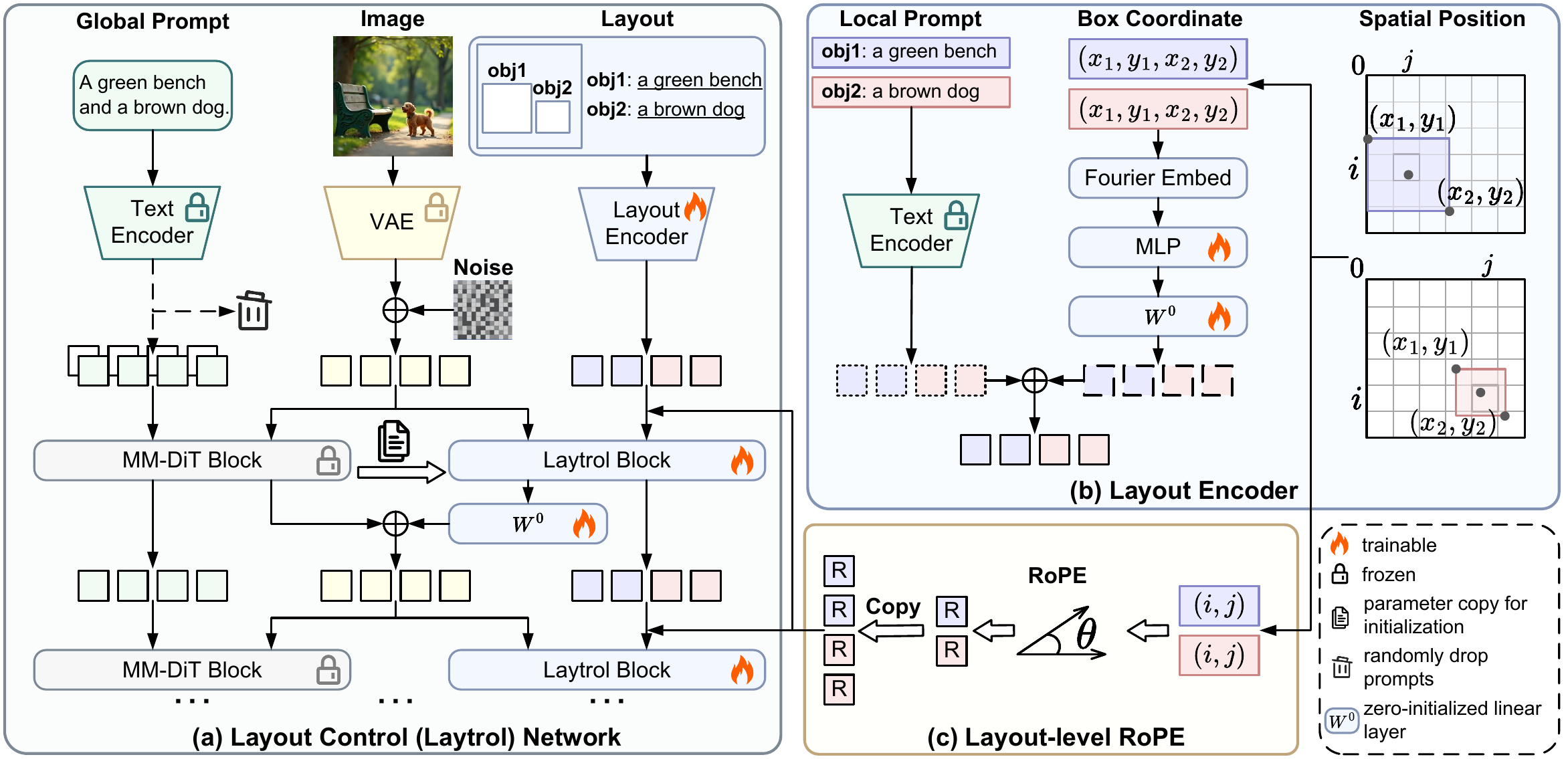}
    \caption{\textbf{Overview of Layout Control (Laytrol) pipeline}. 
    \textbf{(a)} Laytrol blocks inherit vision-language pre-trained knowledge from DiT via parameter copying, which facilitates learning layout-conditioned control.
    Setting the global prompt tokens to null tokens encourages the model to focus more on the layout tokens during training. 
    \textbf{(b)} The layout encoder is initialized as a pure text encoder with a zero-initialized projection layer.
    \textbf{(c)} The coordinates of the patch containing the bounding box center are used for applying RoPE to the layout tokens.
    }
    \label{fig_network}
\end{figure*}

\begin{figure*}[t]
    \centering
    \includegraphics[width=\linewidth]{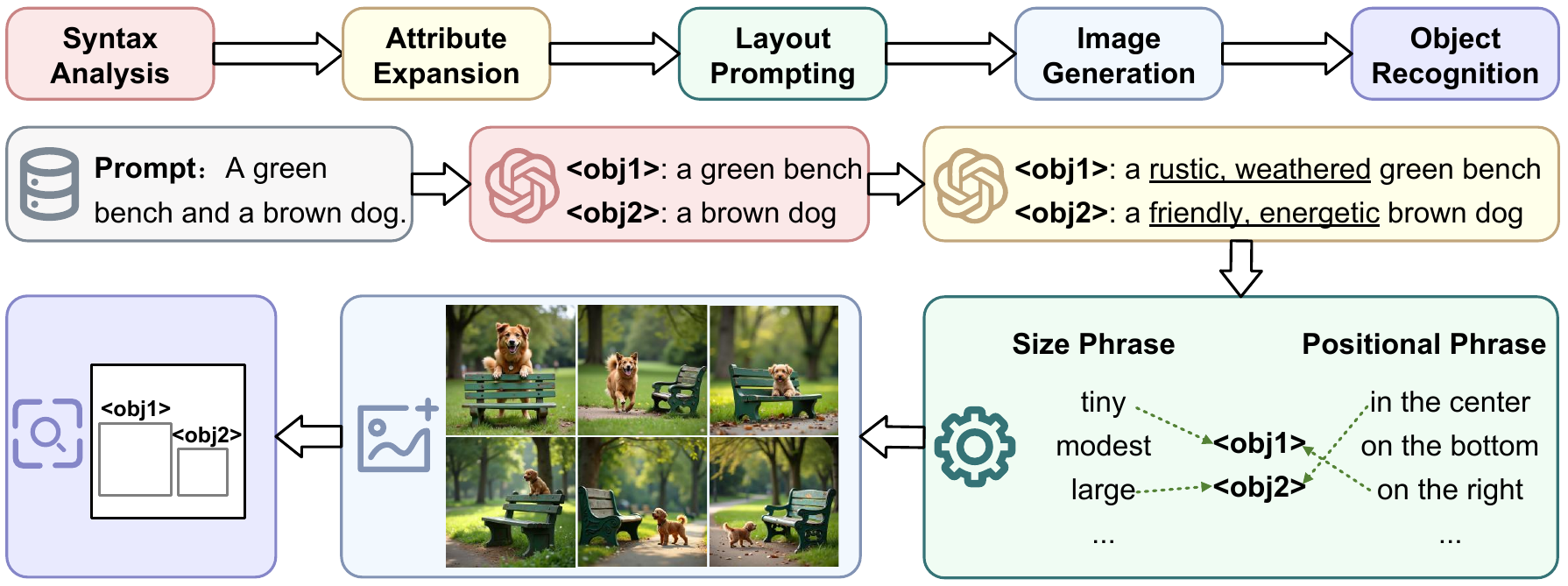}
    \caption{The pipeline for constructing the LaySyn dataset.
    }
    \label{fig_dataset}
\end{figure*}
\subsection{Layout Control (Laytrol) Network}
For U-Net-based diffusion models,  Previous works \cite{GLIGEN, MIGC} encode layout condition $\mathcal{L}$ using Eq. (\ref{eqn_GLIGEN}) and integrate it into the model via inserted cross-attention layers.
For MM-DiT-based models, \cite{CreatiLayout} adopt Eq. (\ref{eqn_GLIGEN}) to derive $\mathbf{C_L}$, which is then mapped to $\mathbf{Q_L}$, $\mathbf{K_L}$, $\mathbf{V_L}$ using newly introduced linear layers, enabling layout control through self-attention.
However, the control modules introduced by existing methods are trained from scratch, limiting their ability to leverage the pretrained knowledge of the base generative model.
Inspired by ControlNet \cite{controlnet}, we propose incorporating copied parameters into the layout control network to mitigate the above limitaition. This approach requires satisfying two initialization conditions \textit{C1} and \textit{C2} as mentioned in Section \ref{sec_intro}.
In the following,  we address \textit{C1} in Layout Condition Encoding and \textit{C2} in Layout Condition Integration.

\subsubsection{Layout Condition Encoding}
The input to ControlNet is formulated as $\mathbf{X}+\mathbf{W^0}\times\mathbf{C_I}$, where $\mathbf{X}$ represents the input image tokens, $\mathbf{W^0}$ is a zero-initialized linear layer, and $\mathbf{C_I}$ denotes the condition image tokens, such as those derived from a depth map or Canny edge map.
By employing a zero-initialized layer with additive fusion, the initial input remains equivalent to $\mathbf{X}$, thereby satisfying Condition $\textit{C1}$.
In contrast to image conditions, layout conditions consist of multiple local prompts $p_i$ paired with their corresponding spatial positions $\mathbf{b}_i$, whose token structure differs significantly from that of $\mathbf{X}$.
This \textbf{feature heterogeneity} prohibits direct addition between the input image tokens $\mathbf{X}$ and the layout condition tokens $\mathbf{C_L}$.

In MM-DiT, the input comprises both image tokens and text tokens. 
The image tokens $\mathbf{X}$ and the local prompt tokens $p_i$ fall within the input domain of MM-DiT, whereas the spatial position tokens $\mathbf{b}_i$ lie outside the input domain.
Consequently, initializing position tokens $\mathbf{b}_i$ to zero is a reasonable choice.
Moreover, due to the intrinsic correspondence between $p_i$ and $\mathbf{b}_i$, it is essential to perform feature fusion between them.

Based on the above considerations, the layout condition tokens $\mathbf{C_L}^i \in \mathbb{R}^{l \times d}$ for each entity $\mathbf{e}_i$ are encoded as
\begin{equation} \label{eqn_C_L_i}
        \mathbf{C_L}^i = \mathrm{T5}(p_i) + \mathbf{W^0} \times \mathrm{MLP}(\mathrm{Fourier}(\mathbf{b}_i)),
\end{equation}
where $\mathrm{T5}$ represents T5 text encoder \cite{T5}.
Prior to the addition, the spatial position tokens are repeated $l$ times to match the length of the local prompt tokens.
All layout condition tokens $\mathbf{C_L}^i$ are then concatenated as
\begin{equation} \label{eqn_C_L}
    \mathbf{C_L} = \bigoplus_{i=1}^{N}\mathbf{C_L}^i \in \mathbb{R}^{(N \cdot l)\times d}.
\end{equation}
The detailed encoding process is illustrated in Figure \ref{fig_network}(b).
 
According Eq. \ref{eqn_C_L_i}, $\mathbf{C_L}^i$ is initially equal to $\mathrm{T5}(p_i)$ at the beginning of training, indicating that $\mathbf{C_L}^i$ represents pure text tokens.
Feeding $\mathbf{C_L}$ into the text branch of MM-DiT can effectively activates the copied parameters, thereby satisfying condition \textit{C1}.
During training, the linear layer $\mathbf{W^0}$ becomes non-zero and gradually injects spatial position features into $\mathbf{C_L}$.
Consequently, $\mathbf{C_L}$ evolves from pure text tokens into layout-aware tokens that integrate both local prompts and spatial positional information.

\subsubsection{Layout Condition Integration}
A standard MM-DiT computes attention over both image and text tokens as follows:
\begin{equation} \label{eqn_dit}
        \mathbf{X_T}',\mathbf{C_T}' = \mathrm{DiT}(\mathbf{X},\mathbf{C_T};\Theta).
\end{equation}
where  $\Theta$ denotes the parameters of the DiT block.

As shown in Figure \ref{fig_network}(a), Laytrol blocks share the same architecture as MM-DiT blocks. For each Laytrol block, its parameters $\Theta_c$ are initialized as a trainable copy of the corresponding MM-DiT parameters $\Theta$.
Given image tokens $\mathbf{X}$ and layout condition tokens $\mathbf{C_L}$ as input, Laytrol block outputs the corresponding transformed tokens:
\begin{equation} \label{eqn_laytrol}
        \mathbf{X_L}',\mathbf{C_L}' = \mathrm{DiT}(\mathbf{X},\mathbf{C_L};\Theta_c).
\end{equation}
In this process, $\mathbf{C_L}$ guides $\mathbf{X}$ to ensure compliance with the layout condition.
During the fusion of $\mathbf{X_T}'$ and $\mathbf{X_L}'$, a zero-initialized linear layer $\mathbf{W}^0$ is added to satisfy condition \textit{C2}:
\begin{equation} \label{eqn_fusion_X}
        \mathbf{X}' = \mathbf{X_T}' + \mathbf{W}^0 \times \mathbf{X_L}'.
\end{equation}
This initialization scheme ensures that the Laytrol network has no effect on the base model at the start of training. Therefore, the model can generate a standard image based solely on the global prompt.
Finally, $\mathbf{X}'$, $\mathbf{C_T}'$, $\mathbf{C_L}'$ are fed into the next layer of the network.

\subsubsection{Layout-Level RoPE}
Before applying self-attention in MM-DiT, Rotary Position Embedding (RoPE) is added to the tokens as shown in Eq. \ref{eqn_rope}.
For an image token at position index $(i,j)$ in the 2D latent space, the corresponding rotation matrix is denoted as $\mathbf{R}(i,j)$, while all text tokens share a common rotation matrix $\mathbf{R}(0,0)$.
To encode coarse spatial information for layout tokens, each layout token associated with a spatial position $\mathbf{b}_i$ is assigned a rotation matrix based on the position index of the patch containing its center coordinate $(\frac{x_1+x_2}{2},\frac{y_1+y_2}{2})$.
This process is illustrated in Figure\ref{fig_network}(c).
This design encourages image tokens located near to the bounding box to attend more to the corresponding layout tokens during attention computation, thereby enhancing layout control in these regions.

\subsection{Training Objective and Strategy}
\label{subsec_training}
During training, the weights of the base generative model are frozen, while only the parameters of the layout encoder and layout control modules are updated.
The model is optimized using the standard denoising diffusion loss:
\begin{equation} \label{eqn_loss}
        \mathrm{loss} = \mathbb{E}_{\mathbf{x}_0,t,p_g,\mathcal{L},\bm{\epsilon} \sim \mathcal{N}(0,\mathbf{I})}
        \left[ \left\| \bm{\epsilon}-\bm{\epsilon}_\theta(\mathbf{x}_t, t, p_g, \mathcal{L}) \right\|_2^2  \right],
\end{equation}
where $t\in[0,1]$ denotes the time step, $\mathbf{x}_0$ is the original image and $\mathbf{x}_t=(1-t)\mathbf{x}_0+t\bm{\epsilon}$ represents the noisy image.
Following \cite{CreatiLayout}, we incorporate a region-aware loss that increases the loss weight within the bounding box regions by a factor of $\lambda$.

It is observed that the model tends to predict layout-related noise at higher timestep $t$, while at lower timestep $t$ the model focuses on noise associated with fine-grained image details. To emphasize layout-related information, a power-law distribution is employed to assign greater weight to higher $t$:
\begin{equation} \label{sample_t}
        \pi(t;\alpha)=\alpha\cdot t^{\alpha-1}, t\in[0,1],
\end{equation}
where we set $\alpha > 1$ in our experiments.

\subsubsection{Random Prompt Dropping}
Modality competition exists among image tokens, global prompt tokens, and layout tokens \cite{CreatiLayout}.
Due to extensive pretraining on image-text pairs, the model tends to rely predominantly on global prompt tokens during generation, resulting in relatively low attention from image tokens to layout tokens.
To mitigate this issue, the global prompt tokens are replaced with null tokens with a probability $\mathrm{p}_d$, thereby encouraging image tokens to attend more to the layout tokens.

\subsection{LaySyn Dataset Construction}
\label{subsec_dataset}
A suitable dataset for training layout-to-image models should annotate each image with a global prompt, entity bounding boxes, and corresponding local prompts.
Previous works \cite{GLIGEN, MIGC, CreatiLayout} adopt LAION \cite{laion-5b, laion-aesthetics-v2} or COCO \cite{COCO} as training datasets.
However, these datasets exhibit distribution shift relative to the pretraining data of models such as FLUX, leading to a degradation in image quality.
To address this issue, we propose Layout Synthesis (LaySyn) dataset.
Images are generated using FLUX and annotated with their corresponding layouts, with the goal of preserving the model’s original generative capability.
During image generation from prompts, we observe that the model often produces images with certain fixed layouts, resulting in limited layout diversity.  We refer to this phenomenon as \textbf{Layout Bias}.
To mitigate layout bias and promote a more uniform distribution of layouts, we propose layout prompting, as illustrated in Figure \ref{fig_dataset}.
By randomly incorporating phrases that describe position and size into the object-level prompts, the generated image layouts are enriched.

The LaySyn dataset construction pipeline consists of five stages, as illustrated in Figure \ref{fig_dataset}. 
First, we select prompts containing multiple objects from the training sets of T2I-CompBench \cite{t2i-compbench} and DiffusionDB \cite{diffusiondb}.
The objects in prompts are localized using syntax analysis performed by GPT-4o \cite{GPT4}.
To enhance the richness of the prompts, adjectival attributes are added to the object descriptions.
During the layout prompting stage, size-related (e.g., tiny, large) and positional (e.g., on the left, in the center) phrases are randomly added to object references.
Subsequently, images are generated using FLUX, and object locations are annotated by Grounding DINO \cite{GroundingDINO}.
Bounding boxes with high Intersection-over-Union (IoU) scores are suppressed to avoid redundancy.
The final dataset contains approximately 400,000 images.

\section{Experiments}

\begin{figure*}[ht]
    \centering
    \includegraphics[width=\linewidth]{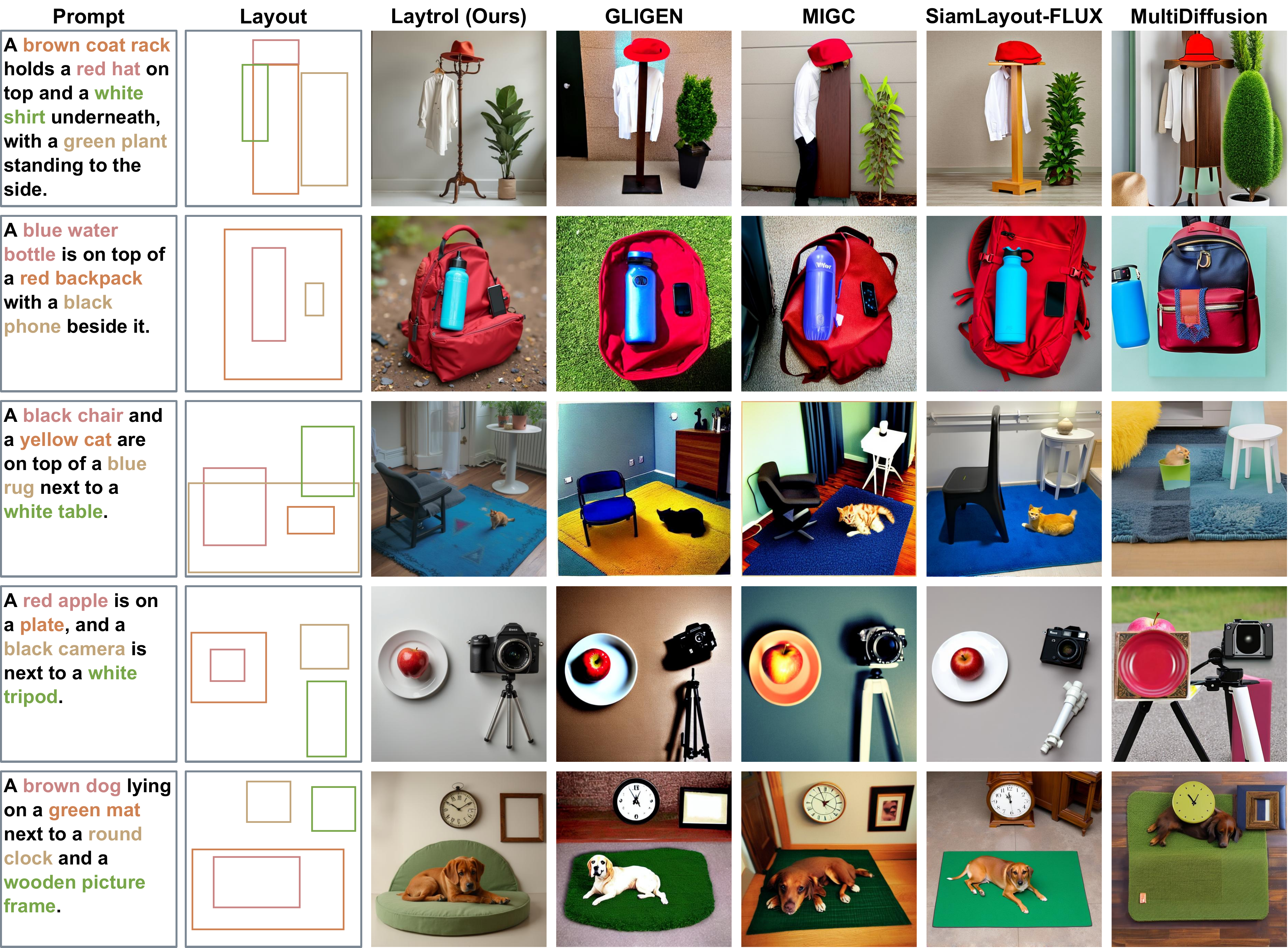}
    \caption{Qualitative comparison with other other methods.
    From the results, Laytrol shows better performance in terms of stylistic consistency, layout realism and object aesthetics.
    }
    \label{fig_visual2}
\end{figure*}

\begin{table*}[t]
    \centering
        \resizebox{1.0\linewidth}{!}{ 
        \begin{tabular}
        {lcccccc}
        \toprule
        \multicolumn{1}{c}
        {\multirow{2}{*}{\bf Model}} & \multicolumn{3}{c}{\bf Attribute Binding } & \multicolumn{2}{c}{\bf Object Relationship} & \multirow{2}{*}{\bf Complex$\uparrow$}
        \\
        \cmidrule(lr){2-4}\cmidrule(lr){5-6}
        &
        {\bf Color $\uparrow$ } &
        {\bf Shape$\uparrow$} &
        {\bf Texture$\uparrow$} &
        {\bf Spatial$\uparrow$} &
        {\bf Non-Spatial$\uparrow$} &
        \\
        \midrule
        Stable Diffusion v1.4 \cite{LDM} & 37.65 & 35.76 & 41.56 & 12.46 & 30.79 & 30.80 \\
        Stable Diffusion v2 \cite{LDM}  & 50.65 & 42.21 & 49.22 & 13.42 & 31.27 & 33.86 \\
        Stable Diffusion XL \cite{LDM}  & 58.79 & 46.87 & 52.99 & 21.33 & 31.19 & 32.37 \\
        FLUX.1 \cite{FLUX}              & 74.07 & 57.18 & 69.22 & 28.63 & 31.27 & 37.03 \\
        \midrule
        GLIGEN \cite{GLIGEN}                & 34.00 & 34.70 & 49.16 & 33.22 & 30.39 & 27.96 \\
            MIGC  \cite{MIGC}               & 65.34 & 45.99 & 60.78 & 36.39 & 29.80 & 33.77 \\
        MultiDiffusion \cite{multidiffusion}& 65.14 & 55.37 & 66.83 & 26.92 & 26.92 & 36.39 \\
        SiamLayout-FLUX \cite{CreatiLayout} & 76.63 & 52.21 & 65.35 & 35.84 & \textbf{31.27} & 36.78 \\
        \midrule
        \textbf{Laytrol (Ours)} & \textbf{80.65} & \textbf{57.70} & \textbf{70.69} & \textbf{47.40} & 30.40 & \textbf{40.36} \\
        \bottomrule
        \end{tabular}
    }
    \caption{Quantitative comparison on T2I-CompBench.} 
    \label{tab_t2i}
\end{table*}

\begin{table}[t]
    \centering
    \begin{tabular}{@{}l@{\hspace{9pt}}c@{\hspace{9pt}}c@{\hspace{9pt}}c@{\hspace{9pt}}c@{\hspace{9pt}}c@{}}
        \toprule
         \bf Model & 
         \bf FID $\downarrow$ & 
         \bf IS $\uparrow$     & 
         \bf mIoU $\uparrow$& 
         \bf AP $\uparrow$& 
         \bf SSIM  $\uparrow$     \\
         \midrule
         GLIGEN&  39.85&  23.94&  79.71
&  68.92
& 16.18            \\
         MIGC&  39.25&  26.58&  77.64
&  65.11
& 20.43            \\
         MultiDiff&  56.71&  20.88&  42.94
&  25.59
& 20.46  \\
         SiamLayout&  36.66&  \textbf{26.81}
&  70.09
&  56.62
& 26.66      \\
         \midrule
         \bf Laytrol & \textbf{34.34} & 26.39 & \textbf{80.08} & \textbf{70.11} & \textbf{27.13} \\
         \bottomrule
    \end{tabular}
    \caption{Quantitative comparison on COCO 2017 dataset.}
    \label{tab_coco}
\end{table}

\begin{table}[t]
    \centering
    \begin{tabular}{cccccc}
        \toprule
         \bf Interval & 
         \bf FID $\downarrow$ & 
         \bf IS $\uparrow$     & 
         \bf mIoU $\uparrow$& 
         \bf AP $\uparrow$& 
         \bf SSIM  $\uparrow$     \\
         \midrule
         1&  35.38&  26.33&  76.75&  64.11& 27.91\\
         2&  37.56&  24.73&  73.54&  59.21& 27.60\\
         4&  39.22&  24.66&  72.37&  57.63& 27.32
\\
         6&  42.48&  23.31&  72.16&  57.44& 27.10\\
        \bottomrule
    \end{tabular}
    \caption{Quantitative experiments on the number of Laytrol blocks. Interval refers to the number of MM-DiT blocks controlled by each Laytrol block.}
    \label{tab_interval}
\end{table}

\begin{table}[t]
    \centering
    \begin{tabular}{@{}ccc|ccc@{}}
        \toprule
         \bf P-Copy & 
         \bf L-RoPE & 
         \bf P-Drop & 
         \bf FID  $\downarrow$ &
         \bf mIoU $\uparrow$   & 
         \bf AP $\uparrow$  \\
         \midrule
          \ding{55}& \ding{55} & \ding{55} &  38.22&  64.92& 51.78
\\
          \ding{51} & \ding{55} & \ding{55} &  35.65&  75.68& 63.31
\\
          \ding{55} & \ding{51} & \ding{55} &  37.20&  67.02& 52.15
\\
          \ding{55} & \ding{55} & \ding{51} &  36.70&  71.28& 56.71
\\
          \ding{51}& \ding{51} & \ding{51} &  \textbf{35.38}&  \textbf{76.75}& \textbf{64.11}\\
         \bottomrule
    \end{tabular}
    \caption{Ablation on COCO 2017 dataset.
    P-Copy, L-RoPE, and P-Drop represent parameter copy, layout-level RoPE, and random prompt dropping, respectively.
    }
    \label{tab_ablation}
\end{table}

\subsection{Experimental Details}

\subsubsection{Training settings}
We train the model on two datasets: the proposed LaySyn dataset and COCO 2017 \cite{COCO}.
FLUX.1-dev is employed as the base generative model.
The training employs the AdamW optimizer with a cosine annealing learning rate schedule, an initial learning rate of 3e-5, and a weight decay of 0.01.
A linear warm-up is applied during the first 1,000 iterations.
The model is trained with a batch size of 64 for 500,000 iterations using DeepSpeed ZeRO Stage 3 on 8 NVIDIA H100 GPUs.
We adopt bf16-mixed precision training \cite{bfloat16} to improve computational efficiency, and apply gradient checkpointing to reduce GPU memory consumption.
The image resolution is $512\times512$.
The region-aware loss weight $\lambda$ is set to 2; the exponent $\alpha$ of the sampling distribution is 1.4; and the prompt dropping probability $p_d$ is 0.5.

\subsubsection{Evaluation settings}
We evaluate our model trained on LaySyn using T2I-Compbench \cite{t2i-compbench}.
This benchmark assesses attribute binding, spatial relationships, and non-spatial relationships.
Before image generation, we generate a layout from each prompt using GPT-4o, following a procedure similar to Section \ref{subsec_dataset}.
Our proposed Laytrol is compared with pretrained text-to-image models including Stable Diffusion \cite{LDM} and FLUX \cite{FLUX}, as well as layout-to-image models including GLIGEN \cite{GLIGEN}, MIGC \cite{MIGC}, MultiDiffusion \cite{multidiffusion} and SiamLayout \cite{CreatiLayout}.
For models trained on COCO, we conduct evaluations using FID \cite{FID}, IS \cite{IS}, mIoU, AP and SSIM\cite{SSIM}.
Following prior work \cite{MIGC}, we employ Grounding-DINO \cite{GroundingDINO} to detect each object and compute the maximum IoU between the detected boxes and the corresponding ground truth box.
The baseline models used for comparison are the aforementioned layout-to-image methods.

\subsection{Qualitative Results}
As shown in Figure \ref{fig_visual2}, the images generated by Laytrol exhibit higher stylistic consistency compared to those produced by GLIGEN and MIGC, which often resemble disjointedly objects and lack scene coherence.
Moreover, Laytrol achieves more realistic layouts. 
For instance, in the first row, Laytrol generates a shirt hanging from a coat rack using a hanger, an element that is absent in other models and leads to more plausible interaction between objects.
In the second row, the water bottle and phone are realistically positioned on the backpack, reflecting a strong adherence to physical plausibility.
In addition, objects generated by Laytrol display superior aesthetic qualities in terms of color, texture and lighting, further validating its ability to preserve pretrained knowledge.

\subsection{Quantitative Results}
\subsubsection{T2I-CompBench}
In Table \ref{tab_t2i}, we evaluate Laytrol and other methods on T2I-CompBench, a comprehensive benchmark for open-world compositional image generation.
For attribute binding, Laytrol achieves moderate improvement in color, texture, and shape.
Regarding spatial relationship, Laytrol demonstrates the most significant improvement, indicating that the generated images closely follow the given layouts and accurately reflect the positional relationships among different objects.

\subsubsection{COCO 2017}
In Table \ref{tab_coco}, we evaluate Laytrol and other layout-to-image methods on COCO-2017 dataset.
Laytrol achieves the best performance in terms of mIoU and AP metrics, demonstrating its superior spatial control capability. Meanwhile, Laytrol also has better results on FID and SSIM, indicating that the images generated by Laytrol exhibit higher overall quality.

\subsection{Laytrol Block Scaling Analysis}
To improve parameter efficiency and reduce inference cost, we scale down the number of Laytrol blocks by adjusting the interval, which specifies how many MM-DiT blocks are controlled by each Laytrol block.
To ease the computational burden, we reduce the number of training iterations to 200,000 in these experiments.
Detailed experimental results are presented in Table \ref{tab_interval}.
Although reducing the number of control blocks leads to a slight decline in mIoU and AP, the layout accuracy remains reasonably preserved.

\subsection{Ablation Study}
We present ablations on the components of Laytrol in Table \ref{tab_ablation}.
For all experiments, the number of training iterations is set to 200,000.
The table demonstrates the effectiveness of the three components: parameter copy, layout-level RoPE, and random prompt dropping.
In these components, parameter copy contributes the most significant improvement to the performance.

\section{Conclusion}
In this work, we propose Laytrol, a layout-to-image generation method that preserves pretrained knowledge through parameter copying.
To ensure effective training, we design an initialization scheme that activates the copied parameters and maintains training stability.
In addition, we introduce the LaySyn dataset to alleviate the domain shift from the pretraining data of the base generative model.
Our work can be further improved by enhancing the diversity of the dataset distribution and integrating multiple types of control conditions.

\bibliography{aaai2026}

\clearpage

\appendix

%
\section{Details of LaySyn Dataset}  \label{appendix_dataset}
To mitigate the distribution shift between the layout-to-image dataset and the pretraining data of the base generative model, we propose Layout Synthesis (LaySyn) dataset.
LaySyn utilizes the base generative model FLUX \cite{FLUX} to synthesize images and employs Grounding DINO \cite{GroundingDINO} to annotate their corresponding layouts.

The LaySyn dataset construction pipeline consists of five stages.
First, we select prompts containing multiple objects from the training sets of T2I-CompBench \cite{t2i-compbench} and DiffusionDB \cite{diffusiondb}.
The mentioned objects are then localized using syntactic analysis conducted by GPT-4o \cite{GPT4}.
To enhance the descriptive richness of the prompts, adjectival attributes are added to the object descriptions.
The following is a reference example used to guide GPT in both syntactic analysis and attribute expansion:

\begin{tcolorbox}[colback=gray!5!white,colframe=gray!80]
\textbf{Caption}: A red teddy bear is holding two balloons of different colors.

\textbf{Objects}: $<$obj\_1$>$ [a red teddy bear] $<$obj\_2$>$ [balloon 1] $<$obj\_3$>$ [balloon 2]

\textbf{Attribute-based Objects}: $<$att\_1$>$ [a soft, plush red teddy bear] $<$att\_2$>$ [a bright blue balloon] $<$att\_3$>$ [a vibrant yellow balloon]

\textbf{Recaptioning}: $<$att\_1$>$ [A soft, plush red teddy bear] $<$oth$>$ [is holding] $<$att\_2$>$ [a bright blue balloon] $<$oth$>$ [and] $<$att\_3$>$ [a vibrant yellow balloon] $<$oth$>$ [in a cheerful setting filled with colorful decorations].

\end{tcolorbox}

\begin{figure*}
    \centering
    \includegraphics[width=\linewidth]{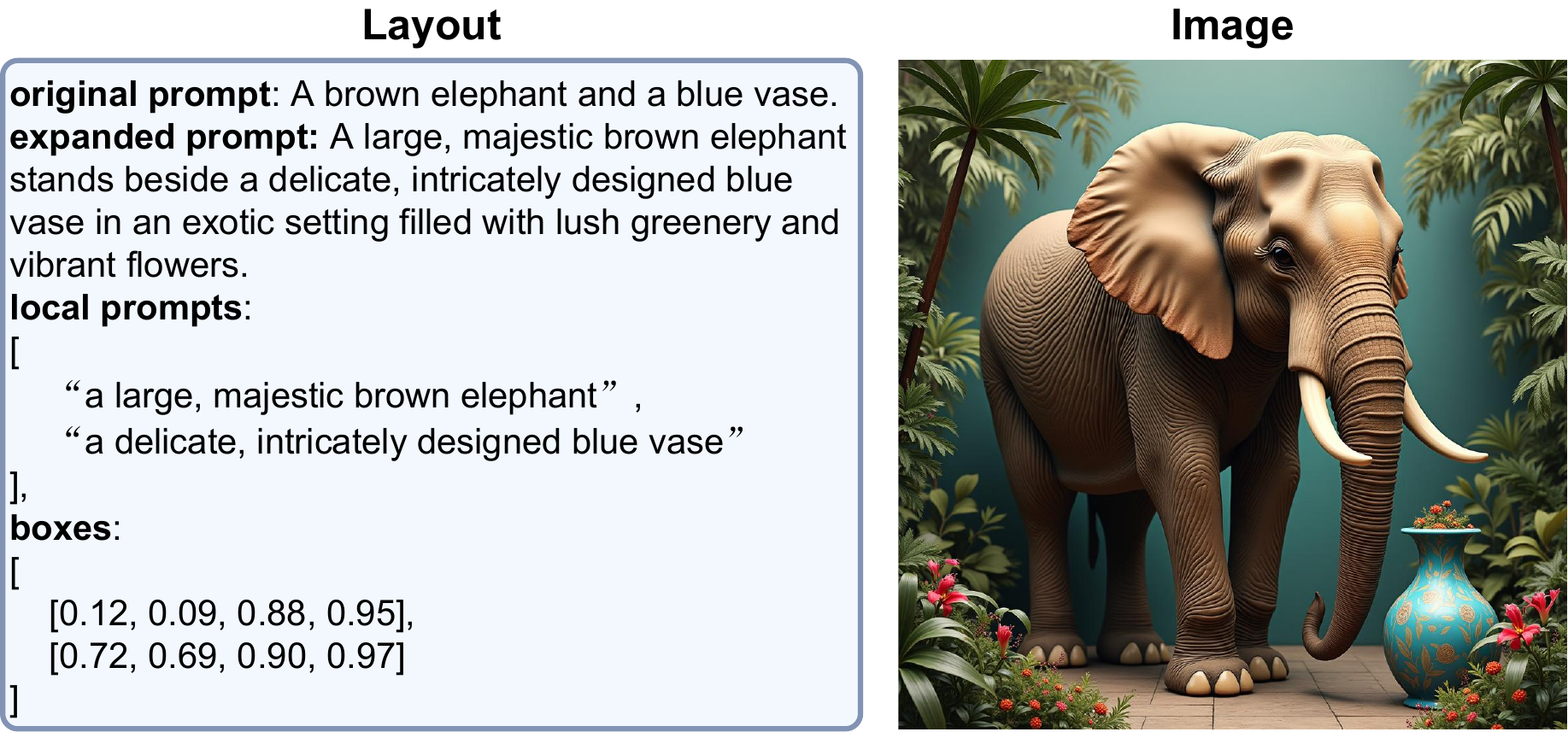}
    \caption{An example data instance from LaySyn dataset.
    }
    \label{fig_dataset_example}
\end{figure*}

During the layout prompting stage, size-related and positional phrases are randomly inserted into object references.
The specific phrases used for layout prompting are listed in Table \ref{tab_layout_prompt}.
After image generation and object recognition, the final images and their corresponding layout annotations are obtained. 
An illustrative example is provided in Figure \ref{fig_dataset_example}.

\section{Additional Results}  \label{appendix_results}
\subsection{Inference Efficiency Analysis}
To further investigate the trade-off between inference efficiency and control capability, we vary the number of controlled DiT blocks per Laytrol block, denoted as $n$. Specifically, a single Laytrol-$n$ block controls $n$ DiT blocks, thus introducing only approximately $1/n$ additional parameters of the base model. 
We conduct ablations on $n\in{1,2,6}$ and evaluate the inference efficiency under the same default generation settings. As shown in Table \ref{tab_inference_costs}, when increasing $n$, the introduced computational and memory decreases significantly, , while still maintaining most controllability performance as shown in Table \ref{tab_interval}. These results demonstrate that Laytrol achieves an adjustable efficiency–controllability balance suitable for practical deployment.

\subsection{Visual Quality Evaluation}
Beyond automatic metrics, we further conduct a comprehensive visual quality evaluation involving 30 human participants and LLM. The evaluation covers three dimensions of aesthetic, realism, and semantic coherence. Each criterion is scored from 0 to 5. The results in Table \ref{tab_human_gpt_eval} demonstrate that Laytrol consistently achieves higher scores than SiamLayout across the three evaluation dimensions.

\subsection{Visualization}
\begin{figure*}
    \centering
    \includegraphics[width=\linewidth]{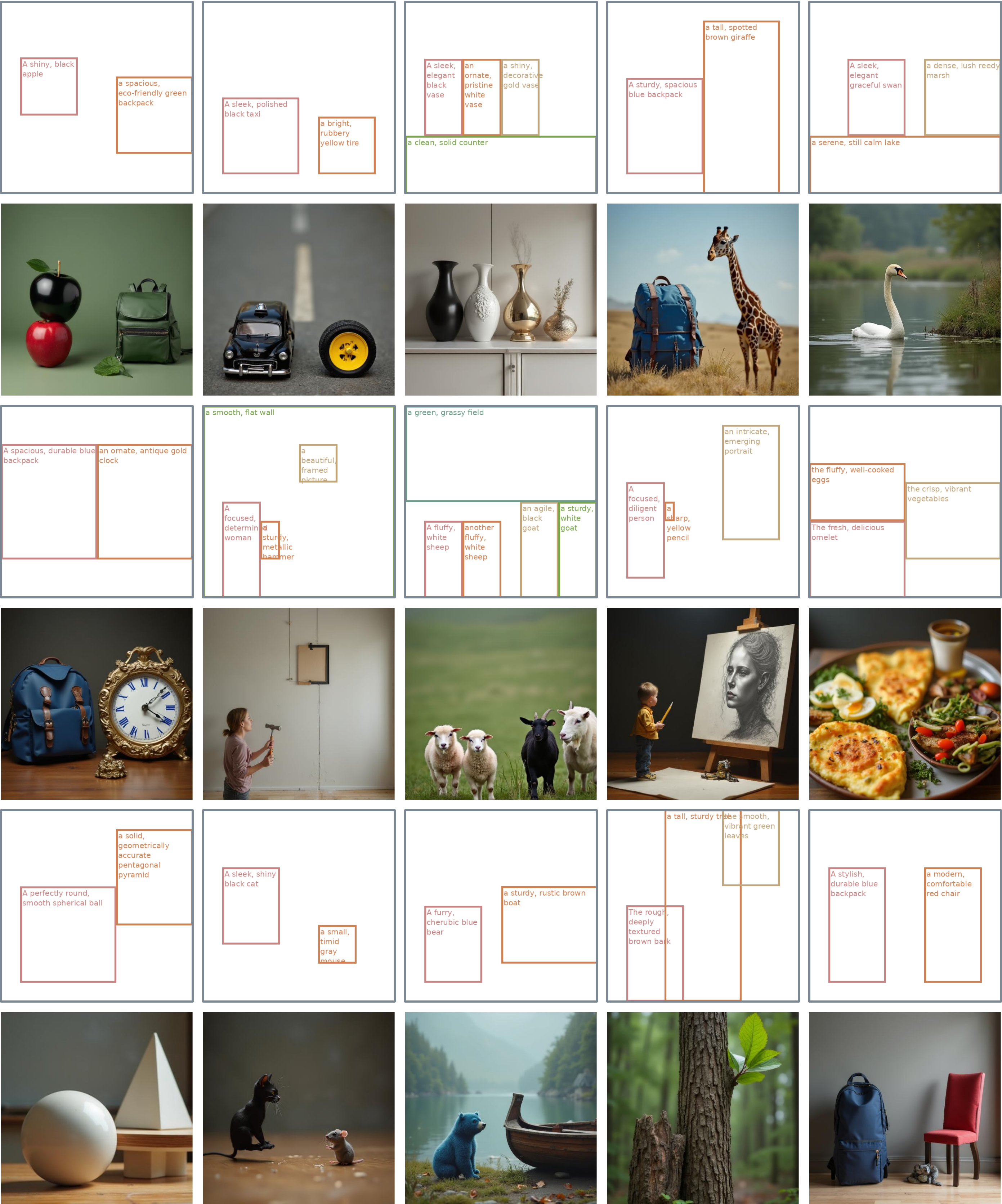}
    \caption{Layout-to-image generation results of Laytrol on T2I-CompBench.
    }
    \label{fig_t2i}
\end{figure*}

\begin{figure*}
    \centering
    \includegraphics[width=\linewidth]{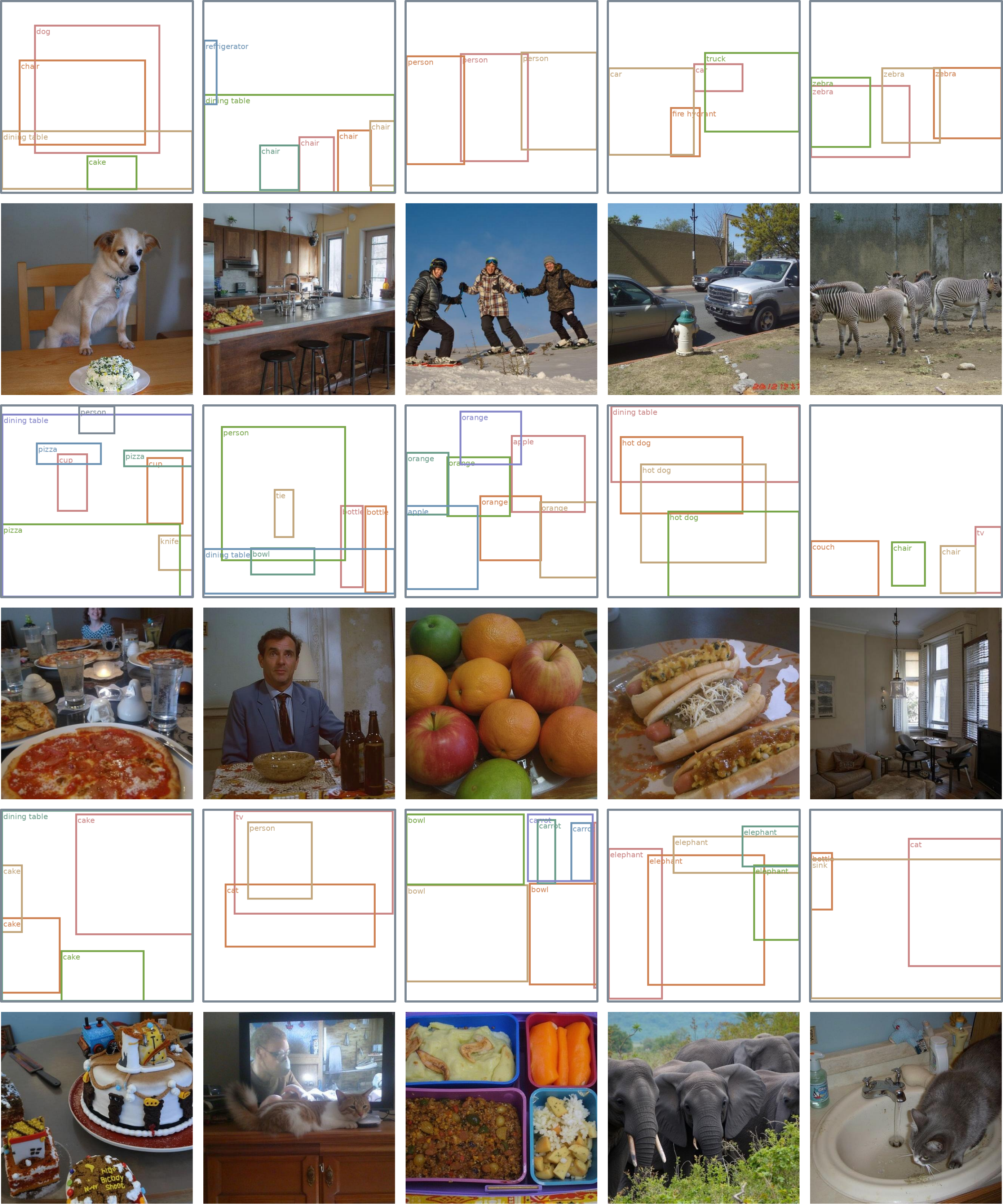}
    \caption{Layout-to-image generation results of Laytrol on COCO 2017 dataset.
    }
    \label{fig_coco}
\end{figure*}

We provide layout-to-image visualization results on the T2I-CompBench \cite{t2i-compbench} and COCO-2017 \cite{COCO} dataset, as shown in Figure \ref{fig_t2i} and Figure \ref{fig_coco}.
For T2I-CompBench, the layouts are generated using GPT-4o, while for COCO-2017, they are derived from annotation labels and box coordinates.
\begin{table}
    \centering
    \begin{tabular}{cc}
        \toprule
         \textbf{Layout Prompting Type} & \textbf{Phrase}\\
         \midrule
         \multirow{5}{*}{positional prompt}& on the top of image \\
         & on the bottom of image \\  
         & on the left of image  \\
         & on the right of image  \\
         & in the center of image  \\
         \midrule
         \multirow{5}{*}{size prompt}& tiny\\
         &  small \\
         &  modest \\
         &  large \\
         &  huge \\
         \bottomrule
    \end{tabular}
    \caption{Size-related and positional phrases in layout prompting phase.}
    \label{tab_layout_prompt}
\end{table}

\begin{table}
    \centering
    \begin{tabular}{@{}c @{} c c c  c@{}}
        \toprule
        Metric & FLUX & Laytrol-1 & Laytrol-2 & Laytrol-6 \\
        \midrule
        TFLOPs & 7.4 & 15.6 & 11.7 & 9.3 \\
        Latency (s) & 2.9 & 6.4 & 5.1 & 4.3 \\
        Memory (GB) & 32.2 & 55.8 & 44.4 & 36.9 \\
        \bottomrule
    \end{tabular}
    \caption{Inference Costs of Different Models Under Default Settings}
    \label{tab_inference_costs}
\end{table}

\begin{table}
    \centering
    \begin{tabular}{c c c c}
        \toprule
        Evaluator & Metric & SiamLayout & Laytrol \\
        \midrule
        \multirow{3}{*}{GPT-4o} 
            & Aesthetics & 3.85 & 4.50 \\
            & Realism & 3.52 & 4.04 \\
            & Coherence & 4.24 & 4.41 \\
        \midrule
        \multirow{3}{*}{Human}
            & Aesthetics & 3.32 & 3.96 \\
            & Realism & 3.58 & 3.72 \\
            & Coherence & 4.09 & 4.24 \\
        \bottomrule
    \end{tabular}
    \caption{Comparison of visual quality evaluation results from human participants and GPT-4o. The evaluation includes aesthetic, realism, and semantic coherence.}
    \label{tab_human_gpt_eval}
\end{table}

\end{document}